\documentclass{article}

\usepackage[preprint, nonatbib]{neurips_2021}

\pdfoutput=1

\usepackage[utf8]{inputenc} 
\usepackage[T1]{fontenc}    
\usepackage{hyperref}       
\usepackage{url}            
\usepackage{booktabs}       
\usepackage{amsfonts}       
\usepackage{nicefrac}       
\usepackage{microtype}      
\usepackage{xcolor}         
\usepackage{amsmath}
\usepackage{caption}
\usepackage{subcaption}
\usepackage[vlined,ruled,commentsnumbered]{algorithm2e}
\usepackage[colorinlistoftodos]{todonotes}

\title{Gradient-Free Neural Network Training via Synaptic-Level Reinforcement Learning}

\author{%
  Aman Bhargava \\
  Division of Engineering Science\\
  University of Toronto\\
  Toronto, Ontario M5S 2E4, Canada \\
  \texttt{aman.bhargava@mail.utoronto.ca} \\
  \And
  Mohammad R. Rezaei \\
  Department of Biomedical Engineering\\
  University of Toronto\\
  Toronto, ON M5S 3G9, Canada \\
  \texttt{mr.rezaei@mail.utoronto.ca} \\
  
  \And
  Milad Lankarany \\ 
  Division of Clinical and Computational Neuroscience, \\ 
  Krembil Brain Institute, University Health Network \\
  
  60 Leonard Ave, Toronto, Ontario M5T 0SB, Canada \\
  \texttt{milad.lankarany@uhnresearch.ca} \\
}

\begin{document}

\maketitle

\begin{abstract}

An ongoing challenge in neural information processing is: how do neurons adjust their connectivity to improve task performance over time (i.e., actualize learning)? It is widely believed that there is a consistent, synaptic-level learning mechanism in specific brain regions, such as the basal ganglia, that actualizes learning. However, the exact nature of this mechanism remains unclear. 
Here we investigate the use of universal synaptic-level algorithms in training connectionist models. Specifically, we propose an algorithm based on reinforcement learning (RL) to generate and apply a simple biologically-inspired synaptic-level learning policy for multi-layer perceptron (MLP) models. In this algorithm, the action space for each MLP synapse consists of a small increase, decrease, or null action on the synapse weight, and the state for each synapse consists of the last two actions and global reward signals. A binary reward signal indicates either an increase or decrease in the model loss between the previous two iterations. This algorithm yields a static synaptic learning policy that enables simultaneous training of over 20,000 parameters (i.e. synapses) and consistent MLP convergence when applied to simulated decision boundary matching and optical character recognition tasks. The static policy is robust as it produces faster and more consistent training relative to the adaptive policy and is agnostic to activation function, network shape, and task. The trained networks yield character recognition performance comparable to identically shaped networks trained with gradient descent. 0 hidden unit character recognition tests yielded an average validation accuracy of 88.28\%, 1.86$\pm$0.47\% higher than the same MLP trained with gradient descent. 32 hidden unit character recognition tests yielded an average validation accuracy of 88.45\%, 1.11$\pm$0.79\% lower than the same MLP trained with gradient descent.
The approach has two significant advantages in comparison to traditional gradient descent-based optimization methods. First, the robustness of our novel method and lack of reliance on gradient computations opens the door to new techniques for training difficult-to-differentiate artificial neural networks such as spiking neural networks (SNNs) and recurrent neural networks (RNNs). 
Second, the method's simplicity provides a unique opportunity for further development of local rule-driven multi-agent connectionist models for machine intelligence analogous to cellular automata.
\end{abstract}

\section{Introduction}

Understanding the method by which biological neurons adjust their connectivity to actualize learning has major consequences for neuroscience and machine learning (ML) alike \cite{unsolved_problems, backprop_brain}. For ML, it would enable highly parallelizable and robust connectionist models that do not rely on traditional gradient descent and backpropagation methods. In neuroscience, it would improve the understanding of fundamental requirements for effective neural computation \cite{neuro_comm_automata}. However, this problem has proven challenging when investigated via analysis and modelling of biological neuron behavior as well as via the generation of computational models of more abstracted connectionist systems \cite{giving_up_control, rl_single_neuron, training_and_inferring, neuron_as_agent}.
Here, we briefly introduce efforts to date from biologically- and computationally-motivated lines of research in this problem.

Biologically-motivated neuroscientific studies in this research area frequently focus on pathways between the cortex, basal ganglia, and thalamus (known as CBGT pathways) \cite{neural_decision, functional_cbgtc}. It is understood that this brain region is deeply involved with action selection in decision making tasks, and that connectivity adjusts over time to maximize a dopaminergic reward signal \cite{cortical_substrate_decisions}. The major unanswered question on the mechanism behind this phenomena is: how is ``credit'' assigned to any one neuron or synapse in order to inform subsequent adjustments in connectivity \cite{cbgt_rubin}? Since the reward signal, action selection, and pathway connectivity changes are spatially and temporally distant from each other, researchers have struggled to find an effective, cohesive solution to this credit assignment problem. Studies are further complicated by the difficulty of simulating large networks of biologically realistic neuron models \cite{realistic_neuron_sim}. Over time, research in this area has broadly converged to the hypothesis that some consistent synaptic-level RL algorithm is employed in these brain regions to actualize learning \cite{cbgt_rubin}.

Computationally-motivated studies on abstracted connectionist models in this area apply RL or related techniques to train neural network models \cite{giving_up_control, training_and_inferring, rl_single_neuron, neuron_as_agent}. Within these studies, neurons are generally framed as RL agents in a partially observable Markov decision process (POMDP) with a reward signal based on network performance on a task. The works differ in terms of their formulation of the agents' state spaces, and some introduce competing reward schemas \cite{giving_up_control} to incentivize biological realism. However, key issues were observed in the art that either place the ``weight'' of the learning task's challenge on existing non-biologically-feasible techniques and/or mis-formulate each neuron's action space. For example, \cite{giving_up_control} models each neuron as a deep Q-learning neural network optimized using gradient descent (policy gradient) methods. The network is then evaluated on OpenAI's Cartpole, a standard task for deep Q-learning neural networks trained using policy gradient methods \cite{openai_gym, barto_RL}. Both \cite{giving_up_control,training_and_inferring} also frame the action space for each neuron as ``fire'' or ``not fire'' rather than as a set of actions to alter synaptic connectivity, an approach that is unsupported by the neuroscientific and modern machine learning literature. \cite{giving_up_control,training_and_inferring,rl_single_neuron,neuron_as_agent} all train separate policies for each neuron/weight rather than training the same policy to be applied by all synapses, as suggested by the neuroscientific literature discussed above. In general, moderate success in network training was achieved in these works.

We proposed the following high-level design changes:

\begin{enumerate}
    \item Frame the fundamental RL agent as the \textit{synapse} rather than the neuron.
    \item Train and apply \textit{the same} synaptic RL policy on all synapses.
    \item Set the \textit{action space} for each synapse to consist of a small increment, a small decrement, and null action on the synapse weight.
    \item Represent synapse \textit{state} as the last $n$ synapse actions and rewards.
    \item Use a \textit{universal binary reward} at each time step representing whether MLP training loss increased or decreased between the most recent two iterations. 
\end{enumerate}

This formulation makes the problem computationally and statistically feasible, particularly for large networks. Training and applying the same synaptic policy for all synapses provides more data to inform the single policy which leads to higher chance of convergence and lower chance of overfitting for a given policy form \cite{aima, barto_RL}. Choosing the synapse as the fundamental reinforcement learning agent also simplifies the action space as biological neurons can have thousands of synapses \cite{thousand_synapses}, and MLPs trained on standard tasks such as optical character recognition can have an arbitrarily high number of synapses per neuron depending on the hidden layer(s) size. A naïve approach for updating neuron connectivity would yield an action space for each neuron with dimensionality on the same order as the layer size.

While the neuroscientific literature suggests a larger and potentially more complicated synapse state space and reward signal that includes additional factors such as neuron/synapse activity \cite{cbgt_rubin,og_stdp}, we find that this simple formulation enables surprisingly effective MLP training, particularly when a static (learned) synaptic learning policy is applied. 
This static policy enables the simultaneous training of over 20,000 parameters and consistently converged for random decision boundary matching and optical character recognition tasks on the notMNIST dataset \cite{notMNIST}. The learned networks also produce character recognition performance on par with identically shaped networks trained with gradient descent. 0 and 32 hidden unit OCR MLPs were trained 5 times using both the proposed synaptic RL method and gradient descent. The 0 hidden unit synaptic RL MLPs yielded a mean final validation accuracy of 88.28$\pm$0.41\% while the 0 hidden unit gradient descent trained MLPs yielded 86.42$\pm$0.22\% final validation accuracy. The 32 hidden unit synaptic RL tests had mean validation accuracy 88.45$\pm$0.6\% while the 32 hidden unit gradient descent tests had 89.56$\pm$0.52\% mean validation accuracy.

\section{Approach}

We seek to understand the extent to which reinforcement learning methods can be applied to the generation of a synaptic-level policy for training MLPs. While MLPs pale in comparison to the complexity of biological neuronal networks, they offer an easily simulated abstract connectionist model that shares some high-level properties with biological neurons \cite{og_perceptron}. The form of the proposed synaptic-level learning algorithm is informed by several hypotheses about biological neurons. Since biological neurons are single cells with limited computational capacity \cite{single_neuron_computation}, we hypothesize that each synapse in a biological neural network applies a relatively simple policy to adjust its connectivity over time. This policy takes into account information including global reward signals (e.g., dopaminergic signals) and the synapse's own past changes in connectivity to inform subsequent changes in connectivity. We suspect that the policy applied by each synapse is roughly the same in a given brain region, and that differences in behaviour and connectivity arise from differences in local information, not from the application of an entirely different policy. This policy results in the maximization of some reward signal over time.

Given these hypotheses, we frame the problem as a POMDP with synapses as the fundamental reinforcement learning (RL) agents. In this work, we apply the temporal difference learning update equations to learn the $Q$-function for training the synapses of MLP neural networks. We test the applicability and generalizability of this method on simulated data and optical character recognition tasks.

\subsection{MLP notation}
The following notation is used for MLPs \cite{goodfellow_dl, aima}:
\begin{itemize}
    \item $d$: Dimensionality of the original data vectors.
    \item $N$: Number of data vectors $\vec x$ and labels $y$ in the training dataset.
    \item $\vec x \in \mathbb R^{d+1}$: Data vectors, assumed to already be augmented with $1$'s in the first index for the bias term. 
    \item $y$: Data vector label. 
    \item $\hat y$: Predicted data vector label using neural network model.
    \item $\vec w^{(i,j)}$: The $i$th neuron's weight vector in layer $j$. 
    \item $\theta(\cdot)$: Activation function (e.g., $\tanh(\cdot), \text{ReLU}(\cdot)$).
    \item $\mathcal L(y, \hat y)$: Loss function (generally Euclidean distance or Cross Entropy loss). 
\end{itemize}

Each scalar component of a neuron weight vector $\vec w^{(i,j)}$ represents a \textit{synapse}. At each non-terminal layer in the network, the activation function $\theta(\cdot)$ is applied element-wise to previous layer output and a bias term is prepended to the resulting vector. During network training, parameters $\vec w^{(i,j)}$ are estimated in order to minimize $\mathcal L(\hat y, y)$.

Datasets $(X,Y)$ are composed of a matrix of data vectors $X$ and a matrix of data labels $Y$ as follows:

\begin{equation}
        X = \begin{bmatrix}
        \vrule & & \vrule \\
        \vec x^{(1)} & \dots & \vec x^{(N)} \\
        \vrule & & \vrule \\
        \end{bmatrix};\,\,\,\,
        Y = \begin{bmatrix}
        \vrule & & \vrule \\
        \vec y^{(1)} & \dots & \vec y^{(N)} \\
        \vrule & & \vrule 
        \end{bmatrix}
\end{equation}

\subsection{Synaptic reinforcement learning}
Gradient-based approaches for inferring MLP parameters $\vec w^{(i,j)}$ require propagation of error signals backward throughout the network and are not easily observed in biological neural networks \cite{backprop_brain}. We frame this problem as a multi-agent RL problem in a POMDP as follows: Each synapse is treated as an RL agent that executes the same policy. This policy maps synapse state to actions (i.e., to alter the synapse weight). The temporal difference update equation is applied to deduce the $Q$-function such that total reward is maximized over time \cite{aima, barto_RL}.

At discrete time steps, we define the following for the proposed synaptic RL model:

\begin{enumerate}
    \item \textbf{Actions}: Each synapse can either increment, decrement, or maintain its value by some small \textit{synaptic learning rate} $\alpha_s>0$.
    \begin{equation}
        a_t \in \mathcal A = \{-\alpha_s, 0, +\alpha_s\}
    \end{equation}
    
    Weight update for a given synapse $k$ in neuron $i$ and layer $j$ is thus given as:
    \begin{equation}
        w_k^{(i,j)} \leftarrow w_k^{(i,j)} + w_k^{(i,j)}.a_t
    \end{equation}
    
    \item \textbf{Reward}: Reward is defined in terms of the training loss at the previous two time steps. Decreased loss is rewarded while increased or equivalent loss is penalized.
    \begin{equation}
    \begin{split}
        R_t &= \text{sign}(\mathcal L_{t-1} - \mathcal L_{t}) \\
            &= {\begin{cases} +1 \text{ if } \mathcal L_{t-1} > \mathcal L_{t} \\ -1 \text{ if } \mathcal L_{t-1} \leq \mathcal L_{t} \end{cases}}
    \end{split}
    \end{equation}
    
    \item \textbf{State}: Each synapse $w_k^{(i,j)}$ ``remembers'' its previous two actions and the previous two global network rewards.
    \begin{equation}
        w_k^{(i,j)}.s_t = \{a_{t-1}, a_{t-2}, R_{t-1}, R_{t-2}\}
    \end{equation}
    
    \item \textbf{Policy}: The $Q$ function gives the expected total reward of a given state-action pair $(s_t, a)$ assuming that all future actions correspond to the highest $Q$-value for the given future state \cite{aima}. The epsilon-greedy synaptic policy $\pi(s_t)$ returns the action $a\in \mathcal A$ with the highest $Q$-value with probability $(1-\epsilon)$. Otherwise, a random action is returned \cite{barto_RL}. 
    \begin{equation}\begin{split}
        Q(s_t, a) &= R_{t+1} + \gamma \max_{a'}Q(s_{t+1}, a') \\
        \pi(s_t) &= \begin{cases}
        \arg\max_{a\in \mathcal A} Q(s_t, a)  & \text{  with}\Pr = 1-\epsilon \\
        \text{random-uniform}(a\in \mathcal A) & \text{  with}\Pr = \epsilon 
        \end{cases}
    \end{split}\end{equation}
    Where $\gamma \in [0,1]$ is the discount for future reward and $\epsilon \in [0,1]$ is the ``exploration probability'' of the policy. If the $Q$ function is accurate, then $\pi(s_t)$ will return the optimal action $a^*_t$ subject to discount factor $\gamma$. Since the state and action spaces in this formulation have low dimensionality, the $Q$ function (and by extension the policy $\pi$) can be implemented as lookup tables of finite size.
\end{enumerate}

\subsection{Training}
In this study, $Q$-value learning is generally a two-fold process where neural network parameters $\vec w$ are trained at the same time as the $Q$ function and, by extension, policy $\pi$. After a policy $\pi$ has been generated, it can also be applied statically. During policy training, the $Q$-values are updated using the following temporal difference learning update equation for $Q$-learning \cite{barto_RL}:
\begin{equation}
    Q(s_t, a_t) \leftarrow Q(s_t, a_t) + \alpha_q[R_{t+1} + \gamma \max_{a'}(Q(s_{t+1}, a') - Q(s_t, a_t))]
\end{equation}
Where $\alpha_q > 0$ is the Q-learning rate. The training pseudocode is as follows:

\begin{algorithm}[H]
\SetKwInOut{Input}{Input}\SetKwInOut{Output}{Output}
\SetAlgoLined
\Input{$Q$ function, dataset $(X,Y)$, $\textit{iterations}$, $\epsilon, \alpha_s, \alpha_q, \gamma$, MLP $\textit{NET}$, boolean $\text{\textit{Train-Policy}}$}
\Output{$Q$ function, updated $\textit{NET}$}
	\BlankLine
	\BlankLine
	$\text{old-loss} \leftarrow \mathcal L(Y, \textit{NET}(X))$\;
	\For{$t=1\dots\text{iterations}$}{
	    \For{each synapse $w$ in $\text{NET}$} {
	        $w.a_t \leftarrow \pi(w.s_t)$\;
	        $w \leftarrow w + w.a_t$\;
	    }
	    $\text{new-loss} \leftarrow \mathcal L(Y, \textit{NET}(X))$\;
		$R_t \leftarrow \text{sign}(\text{old-loss} - \text{new-loss})$\;

		Update each $w.s$ with new global reward $R_t$\;
		
		\If{$\text{Train-Policy}$}{
		    \For{each synapse $w$ in $\text{NET}$}{
		       Apply temporal difference update equation $(7)$ to update $Q$ using $w.s_t$,  $w.a_t$, $w.s_{t-1}$, $w.a_{t-1}$, and $R_t$.
		    }
		}
	}
	
	\textbf{return} $Q,\textit{NET}$
	\caption{Synaptic RL Training Algorithm.}
\end{algorithm}


\section{Experiments \& results}

To determine the efficacy of the proposed synaptic RL technique for MLP training, learning tasks of increasing complexity and dimensionality are employed, starting with random 2D decision boundary matching. Once a static synaptic learning policy is trained on this task, it is re-applied to a different decision boundary task as well as optical character recognition tasks with different network shapes and activation functions. Experiments are run on an AMD Ryzen 9 5900x 4th generation processor.

\begin{figure}
    \centering
    \begin{subfigure}[b]{0.45\textwidth}
        \centering
        \includegraphics[width=\textwidth]{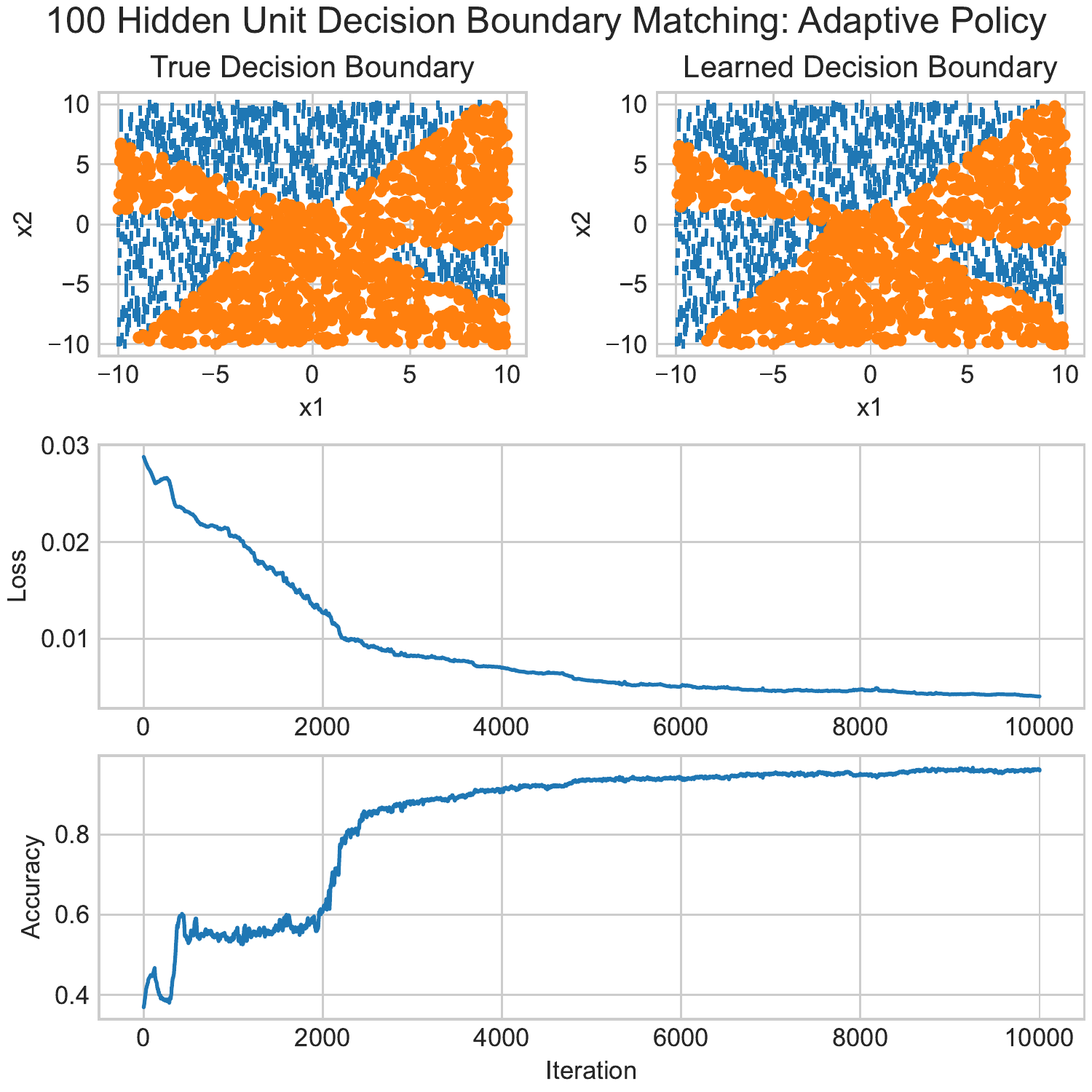}
        \caption{
            Adaptive policy applied to 2D decision boundary matching task. Exploration probability $\epsilon$ is set to $25\%$, Q-learning rate $\alpha_q=0.01$, synaptic learning rate $\alpha_s=0.001$, and future reward discount $\gamma=0.9$. $\tanh$ activation functions were used in both the trained and target networks. Final accuracy is $95.9\%$ while final mean Euclidean loss across the dataset is $0.004023$.
        }
        \label{fig:100HU_adaptive}
    \end{subfigure}
    \hfill
    \begin{subfigure}[b]{0.45\textwidth}
        \centering
        \includegraphics[width=\textwidth]{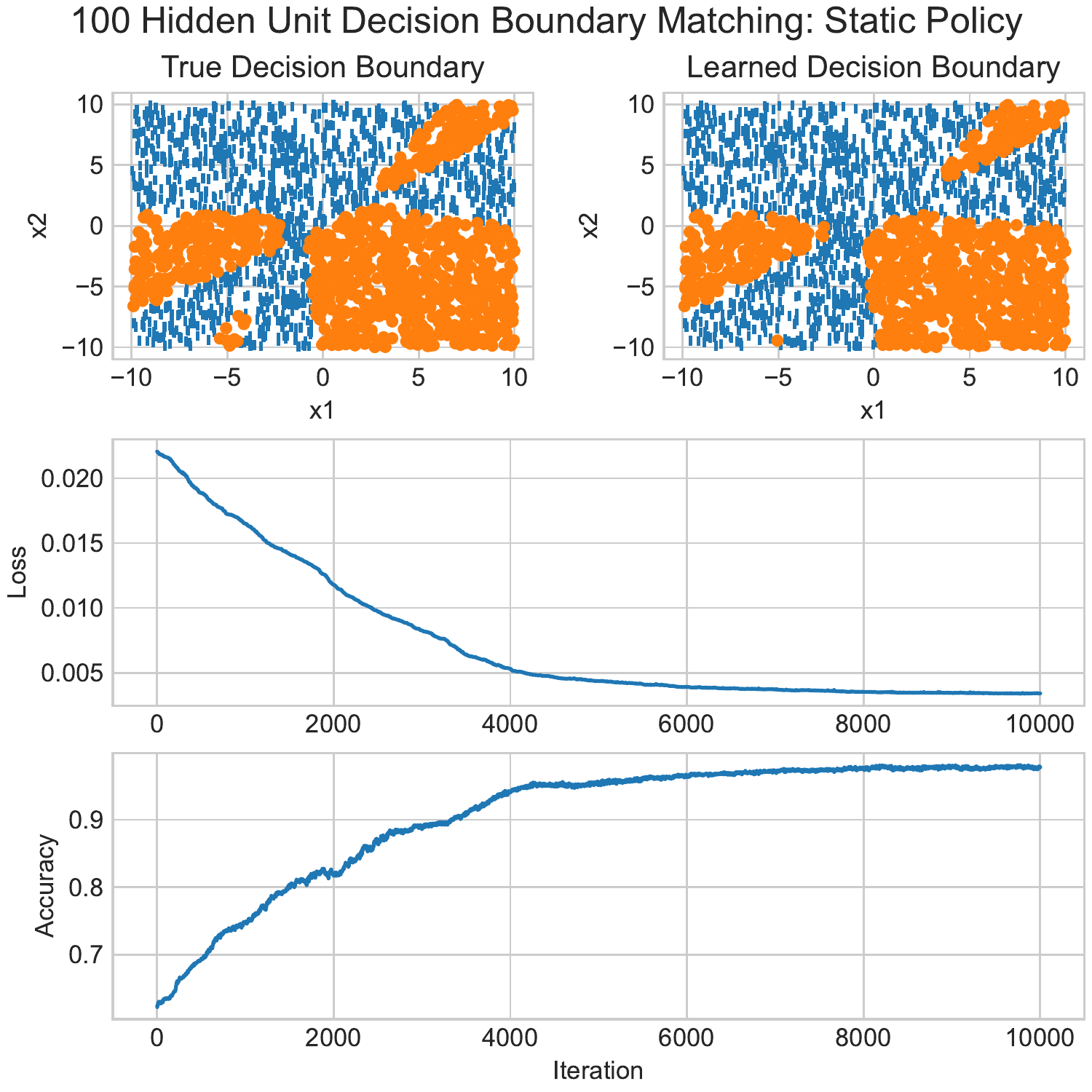}
        \caption{
            Static synaptic policy applied to 2D decision boundary matching task. Exploration probability $\epsilon$ is set to $25\%$, synaptic learning rate $\alpha_s=0.001$, and future reward discount $\gamma=0.9$. $\tanh$ activation functions were used in both the trained and target networks. Final accuracy is $97.7\%$ while final mean Euclidean loss across the dataset is $0.003411$. 
        }
        \label{fig:100HU_static}
    \end{subfigure}
    \caption{Adaptive and static policy applied to decision boundary matching task.}
\end{figure}

\paragraph{MLP binary classification matching on $\mathbb R^2$: } Non-linear 2D decision boundaries are generated by randomly instantiating MLPs with 2 input neurons, 100 hidden units, and 1 output neuron. If the output neuron produces a value greater than zero, the point is classified in the positive class, and vice versa. Synapse weights for the ``target'' MLP are uniformly sampled in the range $[-1, 1]$ and bias terms are set to $0$. Data vector values are uniformly sampled in the range $[-10, 10]$. 2000 data vectors are used in training. The goal of this matching task is to train another MLP of the same size to produce the same classifications as the ``target'' MLP on the random data vectors (i.e., match the decision boundary). Two experiments are highlighted in Figs.~\ref{fig:100HU_adaptive}-\ref{fig:100HU_static}. Fig.~\ref{fig:100HU_adaptive} shows results from simultaneous policy and network training while Fig.~\ref{fig:100HU_static} shows results from applying the policy learned in Fig.~\ref{fig:100HU_adaptive} to a new decision boundary matching problem. All hyperparameters are consistent between the two experiments.

When the synaptic RL policy was trained simultaneously with the network, convergence reliably occurred within 10,000 iterations (Fig.~\ref{fig:100HU_adaptive}). When a static synaptic RL policy from an MLP matching task was re-applied, convergence occurred slightly more quickly and smoothly (Fig.~\ref{fig:100HU_static}).

\paragraph{Optical character recognition on the notMNIST dataset: } The notMNIST dataset is composed of 10 classes of 28x28 greyscale images of typeset characters in a variety of fonts. There are 18,720 labeled images in the dataset, and one-hot encoding is used for the training labels. Cross entropy loss and $\text{ReLU}$ activation functions are used for all trials, and a 75-25\% randomized train-validation split is used for each experiment. Networks with $0$ and $32$ hidden units are trained using both the proposed synaptic RL method and with gradient descent. Hyperparameters for both training methods are manually tuned. The static policy generated in Fig.~\ref{fig:100HU_adaptive} is applied in all synaptic RL OCR trials. All experiments are run 5 times with validation accuracy statistics shown in Table~\ref{table:ocr_results}. Estimated runtime for each individual experiment run, along with the number of trainable parameters, is reported in Table~\ref{table:ocr_results} as well.

\begin{table}
  \caption{notMNIST OCR validation accuracy comparison}
  \label{table:ocr_results}
  \centering
  \begin{tabular}{lllllll}
    \toprule
    \cmidrule(r){1-7}
    \textbf{Experiment}   & Min         & Max        & Mean       & Stdev   & Est. Runtime  & Params    \\
    \midrule
    Syn. RL 32 HU         & 89.12\%     & 89.12\%    &  88.45\%   &  0.60\% & 5.5 hours     & 25,450    \\
    Grad. Desc. 32 HU     & 88.57\%     & 90.00\%    &  89.56\%   &  0.52\% & 20 minutes    & 25,450    \\
    Syn. RL 0 HU          & 87.65\%     & 88.84\%    &  88.28\%   &  0.41\% & 1.5 hours     & 7,850     \\
    Grad. Desc. 0 HU      & 86.11\%     & 86.73\%    &  86.42\%   &  0.22\% & 5 minutes     & 7,850     \\
    \bottomrule
  \end{tabular}
\end{table}

To improve runtime for synaptic-level reinforcement learning experiments, minibatching is used wherein a random subset of the data was re-selected to train on at regular intervals. Batches composed of 5000 training data samples are re-selected every 5000 iterations for all synaptic reinforcement learning experiments. This approach is founded on the premise that prediction error on a sufficiently large subset of the training dataset likely reflects prediction error on the full set (as in stochastic gradient descent) \cite{goodfellow_dl}. Exploration probability $\epsilon=0.1$ and synaptic learning rate $\alpha_s = 0.0001$. For the 32 hidden unit experiments, the learning rate $\alpha_s$ is reduced to $0.00005$ (i.e., reduced by half) at iteration 60,000 to promote network convergence. 32 hidden unit networks are trained for 500,000 iterations and 0 hidden unit networks are trained for 200,000 iterations. 

\begin{figure}
    \centering
    \includegraphics[width=0.65\columnwidth]{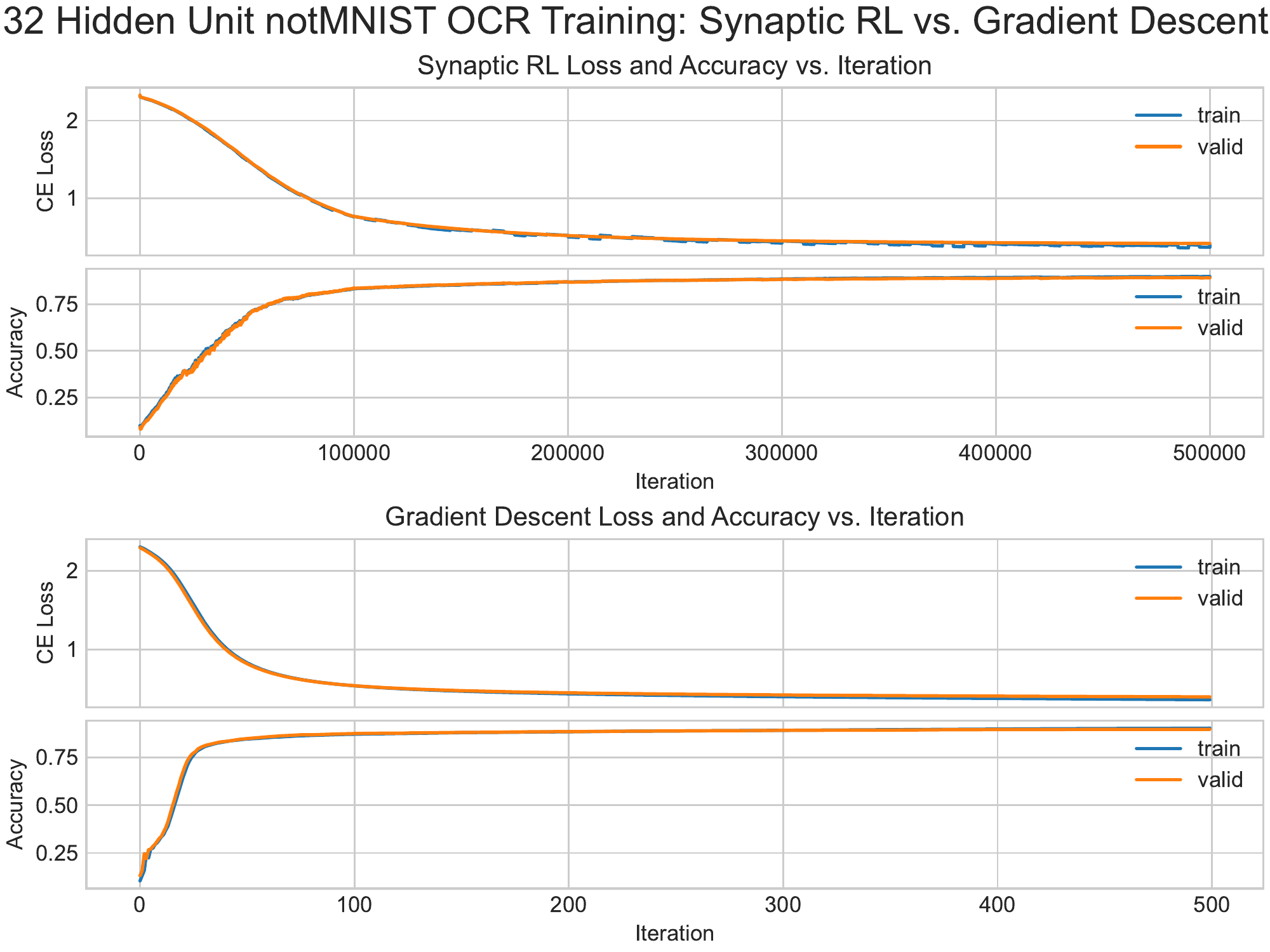}
    \caption{
        Training curves from single trials from the 32 hidden unit synaptic RL and gradient descent experiments. 
    }
    \label{fig:ocr_plots}
\end{figure}

For the gradient descent experiments, batch gradient descent is employed with learning rate $\alpha=0.1$. No regularization, momentum, or mini-batching is used. Duration of training and learning rate is tuned such that the models train to completion (i.e., plateauing validation loss and accuracy). A comparison of training and validation loss and accuracy plots from 32 hidden unit experiments can be seen in Fig~\ref{fig:ocr_plots}.

Taking the difference between mean validation accuracy for each method indicates that the 32 hidden unit synaptic RL method has 1.11$\pm$0.79\% lower validation accuracy than the 32 hidden unit MLPs trained using gradient descent. The 0 hidden unit synaptic RL method has 1.86$\pm$0.47\% higher validation accuracy relative to the same size MLPs trained using gradient descent.  

\section{Discussion}
In this paper, we developed an algorithm to produce and apply a simple, effective synaptic-level learning policy for MLP training. Reapplying the static policy generated by training MLPs on a random decision boundary matching task resulted in networks reliably converging for previously unseen 2-dimensional decision boundaries and optical character recognition tasks. Optical character recognition validation accuracy was also comparable with, if not slightly higher than, neural networks trained using conventional gradient descent learning methods (Table~\ref{table:ocr_results}). Overall, the static synaptic learning policy was robust in the face of new activation functions, new tasks, and new network structures. 
The simplicity, effectiveness, and robustness of this approach for training MLPs supports the hypothesis that simple, universal synaptic-level algorithms can enable learning in large connectionist models. The results from this investigation are particularly salient when compared to previous computational work in this area as the learned, static RL policy was \textit{more} effective in general than the adaptive one.

Of note is the fact that the 0 hidden unit experiments yielded superior validation accuracy (1.86$\pm$0.47\% higher) for the synaptic RL method as compared to gradient descent. We suspect that this may be due to the minibatching approach used for all synaptic RL experiments. This may have helped the synaptic RL method escape local optima that the batch gradient descent method was susceptible to, much like the stochastic gradient descent algorithm upon which the minibatching approach is based \cite{goodfellow_dl}. Further work is warranted to understand this performance difference, including direct comparison of the synaptic RL algorithm with stochastic gradient descent methods.

A key limitation of the algorithm is that the synaptic RL models required many iterations and relatively small synaptic learning rates to converge reliably. The overall time complexity for training MLPs on optical character recognition tasks was higher for the proposed algorithm as compared to gradient descent methods. Since MLPs are easily differentiable, it is logical to rely upon techniques that take advantage of this valuable information to train parameters efficiently. Therefore, we do not propose this algorithm as a replacement for conventional training algorithms, but rather as a demonstration of the effectiveness of simple synaptic-level learning rules in connectionist models. As well, while this study took significant inspiration from neuroscientific literature, MLPs are incredibly simplistic models of the true behaviour of biological neurons \cite{bio_artificial_neuron_comparison}, notably lacking significant time-dependence. Further research is required to determine the extent to which the results obtained are applicable to the credit assignment problem in biological neurons.

Future directions may include investigations on the applicability of this method to other connectionist models such as recurrent neural networks (RNNs), convolutional neural networks (CNNs), and spiking neural networks (SNNs). It is conceivable that the policy generated in this paper could apply directly to RNNs since they are definitionally feedforward neural networks (MLPs) with additional edges that forward propagate to subsequent time steps \cite{rnn_review}. While this complicates backpropagation, the proposed method would remain the same as the only non-local information required in each synapse is the binary reward signal. It is of interest to determine the applicability of the static policy proposed in this work to CNNs as there is substantially more structure imposed on the application of kernels and these kernel ``synapses'' generally coexist with a set of fully connected terminal layers (MLP) \cite{cnn_review, deep_learning}. Since 2D convolutions can be expressed as matrix multiplications with circulant matrices \cite{circ_cnn}, it may be that the proposed algorithm would generalize to the CNN architecture. Applying these findings to SNNs may require significant changes as the large number of iterations required for MLP training would be difficult to achieve for compute-intensive SNNs \cite{og_snn_review}. The neuroscience literature would suggest that time dependent local information features such as pre- and post-synaptic activation timing (as in spike-timing-dependent synaptic plasticity) would be a valuable addition to the state space of each synapse \cite{og_stdp}. Applications in unsupervised learning (e.g., autoencoders) are also of strong interest.

Incorporation of additional information for use in the synaptic reinforcement learning policy on MLPs may further enhance training speed and reliability. Additional testing with different datasets and a more extensive hyperparameter search would also aid in further understanding the broader applicability of the algorithm. Techniques such as momentum \cite{goodfellow_dl} typically applied to gradient descent optimization methods may also be applicable in the design of even more effective synaptic reinforcement learning algorithms. In-depth comparison to works that attempt to create biologically-feasible approximation methods for gradient optimization would also help to shed light on the nature of the proposed algorithm and its similarities to existing methods \cite{grad_est_credit_assig, random_backprop}. 

The algorithm presented in this paper demonstrates the effectiveness of framing the training of MLPs as a multi-agent, single-policy reinforcement learning problem. While gradient based methods are well established for MLPs, our approach removes the requirement for easy differentiation of the network and the loss function. For the future development of machine learning and machine intelligence systems, this is not an insignificant constraint. After all, the primary example of ``true intelligence'' available to us is the brain -- a structure that does not appear to be driven directly by backpropagation-like gradient optimization methods. It is therefore highly valuable to investigate alternative optimization methods for connectionist models.

Furthermore, much of the prior computational work on framing neural network training as a multi-agent reinforcement learning problem struggled with computational efficiency \cite{giving_up_control, rl_single_neuron, training_and_inferring, neuron_as_agent}. Neuron connectivity and policy updates take a long time and are difficult to run at large scale on modern hardware, especially in methods that train different policy for each neuron. The proposed algorithm offers a low memory usage and low complexity single-policy synaptic learning method that shows comparable end-performance to gradient-based methods in training MLPs.

The synapse policy generated and implemented in this work bares resemblance to the cellular automata studied extensively by Wolfram, Conway, Dennett, and others \cite{wolfram_nks, game_of_life, dennett_real_patterns}. One of the major results in Wolfram's \textit{A New Kind of Science} is the fact that complex, irreducible behaviour can result from the application of simple rules \cite{wolfram_nks}. In addition to being complex, the behavior exhibited by the static MLP synapse model in this study is effective in actualizing a high-level goal (MLP training), agnostic of network topology, activation function, and task. While the particular task of MLP training may be ``reducible'' in the sense that gradient computations can enable estimation of near-optimal MLP parameters quickly, Wolfram's work suggests that other potentially useful models for machine intelligence may not have such a ``short cut''.

\section{Broader impact}

We believe that the currently foreseeable consequences of this work will be positive. Robust synaptic-level learning mechanisms for connectionist models open many doors for engineering learning systems, and such mechanisms may also shed light on adjacent neuroscientific problems. Since the main result from this study was that the proposed algorithm yields comparable end-results to standard methods for MLP training with less information, it is unlikely that novel negative societal impacts would arise directly due to this research. 

\section{Acknowledgement}

We thank Emre R. Alca for the use of his computing infrastructure to run the final experiments in this work.


\begin{thebibliography}{10}

\bibitem{unsolved_problems}
R.~{Adolphs}.
\newblock The unsolved problems of neuroscience.
\newblock {\em Trends in Cognitive Sciences}, 19(4):173--175, 2015.

\bibitem{openai_gym}
G.~Brockman, V.~Cheung, L.~Pettersson, J.~Schneider, J.~Schulman, J.~Tang, and
  W.~Zaremba.
\newblock Openai gym, 2016.

\bibitem{notMNIST}
Y.~Bulatov.
\newblock notmnist dataset, September 2011.

\bibitem{og_stdp}
N.~Caporale and Y.~Dan.
\newblock Spike timing--dependent plasticity: a hebbian learning rule.
\newblock {\em Annu. Rev. Neurosci.}, 31:25--46, 2008.

\bibitem{training_and_inferring}
M.~{Chalk}, G.~{Tkacik}, and O.~{Marre}.
\newblock Training and inferring neural network function with multi-agent
  reinforcement learning.
\newblock {\em bioRxiv}, page 598086, 2020.

\bibitem{game_of_life}
J.~Conway.
\newblock The game of life.
\newblock {\em Scientific American}, 223(4):4, 1970.

\bibitem{cortical_substrate_decisions}
N.~D. {Daw}, J.~P. {O'Doherty}, P.~{Dayan}, B.~{Seymour}, and R.~J. {Dolan}.
\newblock Cortical substrates for exploratory decisions in humans.
\newblock {\em Nature}, 441(7095):876--879, 2006.

\bibitem{dennett_real_patterns}
D.~C. Dennett.
\newblock Real patterns.
\newblock {\em The Journal of Philosophy}, 88(1):27--51, 1991.

\bibitem{circ_cnn}
C.~Ding, S.~Liao, Y.~Wang, Z.~Li, N.~Liu, Y.~Zhuo, C.~Wang, X.~Qian, Y.~Bai,
  G.~Yuan, et~al.
\newblock Circnn: accelerating and compressing deep neural networks using
  block-circulant weight matrices.
\newblock In {\em Proceedings of the 50th Annual IEEE/ACM International
  Symposium on Microarchitecture}, pages 395--408, 2017.

\bibitem{realistic_neuron_sim}
E.~D’Angelo, S.~Solinas, J.~Garrido, C.~Casellato, A.~Pedrocchi, J.~Mapelli,
  D.~Gandolfi, and F.~Prestori.
\newblock Realistic modeling of neurons and networks: towards brain simulation.
\newblock {\em Functional neurology}, 28(3):153, 2013.

\bibitem{bio_artificial_neuron_comparison}
O.~Eluyode and D.~T. Akomolafe.
\newblock Comparative study of biological and artificial neural networks.
\newblock {\em European Journal of Applied Engineering and Scientific
  Research}, 2(1):36--46, 2013.

\bibitem{neural_decision}
J.~I. {Gold} and M.~N. {Shadlen}.
\newblock The neural basis of decision making.
\newblock {\em Annual Review of Neuroscience}, 30(1):535--574, 2007.

\bibitem{goodfellow_dl}
I.~Goodfellow, Y.~Bengio, and A.~Courville.
\newblock {\em Deep Learning}.
\newblock MIT Press, 2016.
\newblock \url{http://www.deeplearningbook.org}.

\bibitem{thousand_synapses}
J.~{Hawkins} and S.~{Ahmad}.
\newblock Why neurons have thousands of synapses, a theory of sequence memory
  in neocortex.
\newblock {\em Frontiers in Neural Circuits}, 10:23--23, 2016.

\bibitem{grad_est_credit_assig}
B.~J. {Lansdell}, P.~R. {Prakash}, and K.~P. {Kording}.
\newblock Learning to solve the credit assignment problem.
\newblock {\em arXiv preprint arXiv:1906.00889}, 2019.

\bibitem{deep_learning}
Y.~LeCun, Y.~Bengio, and G.~Hinton.
\newblock Deep learning.
\newblock {\em nature}, 521(7553):436--444, 2015.

\bibitem{random_backprop}
T.~P. {Lillicrap}, D.~{Cownden}, D.~B. {Tweed}, and C.~J. {Akerman}.
\newblock Random synaptic feedback weights support error backpropagation for
  deep learning.
\newblock {\em Nature Communications}, 7(1):13276--13276, 2016.

\bibitem{rnn_review}
Z.~C. Lipton, J.~Berkowitz, and C.~Elkan.
\newblock A critical review of recurrent neural networks for sequence learning.
\newblock {\em arXiv preprint arXiv:1506.00019}, 2015.

\bibitem{single_neuron_computation}
T.~M. McKenna, J.~L. Davis, and S.~F. Zornetzer.
\newblock {\em Single neuron computation}.
\newblock Academic Press, 2014.

\bibitem{neuron_as_agent}
S.~{Ohsawa}, K.~{Akuzawa}, T.~{Matsushima}, G.~{Bezerra}, Y.~{Iwasawa},
  H.~{Kajino}, S.~{Takenaka}, and Y.~{Matsuo}.
\newblock Neuron as an agent.
\newblock In {\em ICLR 2018 : International Conference on Learning
  Representations 2018}, 2018.

\bibitem{giving_up_control}
J.~{Ott}.
\newblock Giving up control: Neurons as reinforcement learning agents.
\newblock {\em arXiv preprint arXiv:2003.11642}, 2020.

\bibitem{functional_cbgtc}
A.~{Parent} and L.-N. {Hazrati}.
\newblock Functional anatomy of the basal ganglia. i. the cortico-basal
  ganglia-thalamo-cortical loop.
\newblock {\em Brain Research Reviews}, 20(1):91--127, 1995.

\bibitem{og_snn_review}
M.~Pfeiffer and T.~Pfeil.
\newblock Deep learning with spiking neurons: opportunities and challenges.
\newblock {\em Frontiers in neuroscience}, 12:774, 2018.

\bibitem{cnn_review}
W.~Rawat and Z.~Wang.
\newblock Deep convolutional neural networks for image classification: A
  comprehensive review.
\newblock {\em Neural computation}, 29(9):2352--2449, 2017.

\bibitem{og_perceptron}
F.~{Rosenblatt}.
\newblock The perceptron: a probabilistic model for information storage and
  organization in the brain.
\newblock {\em Psychological Review}, 65(6):386--408, 1958.

\bibitem{cbgt_rubin}
J.~E. {Rubin}, C.~{Vich}, M.~{Clapp}, K.~{Noneman}, and T.~{Verstynen}.
\newblock The credit assignment problem in cortico‐basal ganglia‐thalamic
  networks: A review, a problem and a possible solution.
\newblock {\em European Journal of Neuroscience}, 53(7):2234--2253, 2021.

\bibitem{aima}
S.~Russell.
\newblock {\em Artificial intelligence : a modern approach}.
\newblock Prentice Hall, Upper Saddle River, New Jersey, 2010.

\bibitem{neuro_comm_automata}
L.~{Su}, R.~{Gomez}, and H.~{Bowman}.
\newblock Analysing neurobiological models using communicating automata.
\newblock {\em Formal Aspects of Computing}, 26(6):1169--1204, 2014.

\bibitem{barto_RL}
R.~S. Sutton and A.~G. Barto.
\newblock {\em Reinforcement learning: An introduction}.
\newblock MIT press, 2018.

\bibitem{rl_single_neuron}
Z.~{Wang} and M.~{Cai}.
\newblock Reinforcement learning applied to single neuron.
\newblock {\em arXiv preprint arXiv:1505.04150}, 2015.

\bibitem{backprop_brain}
J.~C. {Whittington} and R.~{Bogacz}.
\newblock Theories of error back-propagation in the brain.
\newblock {\em Trends in Cognitive Sciences}, 23(3):235--250, 2019.

\bibitem{wolfram_nks}
S.~Wolfram.
\newblock {\em A New Kind of Science}.
\newblock Wolfram Media, 2002.

\end{thebibliography}
\end{document}